\pdfoutput=1

\documentclass[11pt]{article}

\usepackage{acl}

\usepackage{times}
\usepackage{latexsym}

\usepackage[T1]{fontenc}

\usepackage[utf8]{inputenc}

\usepackage{microtype}

\usepackage{times}
\usepackage{latexsym}
\usepackage{graphicx}
\usepackage{bbding}
\usepackage{pifont}
\usepackage{amsfonts,amssymb}
\usepackage{booktabs}
\usepackage{amsmath}
\usepackage{multirow}
\usepackage{bm}
\usepackage{subfigure}
\usepackage{framed}
\DeclareMathOperator*{\argmax}{argmax}
\DeclareMathOperator*{\softmax}{Softmax}

\usepackage{graphics}
\usepackage{caption}
\makeatletter
\def\maketag@@@#1{\hbox{\m@th\normalfont\normalsize#1}}
\makeatother

\usepackage{diagbox}

\usepackage{adjustbox}

\title{Investigating Non-local Features for Neural Constituency Parsing}

\author{
  Leyang Cui$^{\heartsuit \spadesuit}$\thanks{\quad The first two authors contributed equally to this work.}, \ \ Sen Yang$^{\spadesuit *}$, \ \ Yue Zhang$^{\spadesuit \Diamond}$\thanks{\quad Corresponding author.} \\
  $^\heartsuit$Zhejiang University \\
  $^\spadesuit$School of Engineering, Westlake University \\
  $^\Diamond$Institute of Advanced Technology, Westlake Institute for Advanced Study \\
 \texttt{\{cuileyang, zhangyue\}@westlake.edu.cn} \hspace{5pt} \texttt{senyang.stu@gmail.com}
  
  }

\begin{document}
\maketitle
\begin{abstract}

Thanks to the strong representation power of neural encoders, neural chart-based parsers have achieved highly competitive performance by using local features. 
Recently, it has been shown that non-local features in CRF structures lead to improvements. 
In this paper, we investigate injecting non-local features into the training process of a local span-based parser, by predicting constituent $n$-gram non-local patterns and ensuring consistency between non-local patterns and local constituents. 
Results show that our simple method gives better results than the self-attentive parser on both PTB and CTB. 
Besides, our method achieves state-of-the-art BERT-based performance on PTB (95.92 F1) and strong performance on CTB (92.31 F1). 
Our parser also achieves better or competitive performance in multilingual and zero-shot cross-domain settings compared with the baseline.

\end{abstract}

\section{Introduction}

Constituency parsing is a fundamental task in natural language processing, which provides useful information for downstream tasks such as machine translation~\cite{tree-mt}, natural language inference~\cite{nli}, text summarization~\cite{cp-summarization}. Over the recent years, with advance in deep learning and pre-training, neural chart-based constituency parsers~\cite{minimal-span,san-constituency} have achieved highly competitive results on benchmarks like Penn Treebank (PTB) and Penn Chinese Treebank (CTB) by solely using local span prediction.


The above methods take the contextualized representation (e.g., BERT) of a text span as input, and use a local classifier network to calculate the scores of the span being a syntactic constituent, together with its constituent label. For testing, the output layer uses a non-parametric dynamic programming algorithm (e.g., CKY) to find the highest-scoring tree. Without explicitly modeling structured dependencies between different constituents, the methods give competitive results compared to non-local discrete parsers ~\cite{minimal-span,san-constituency}.
One possible explanation for their strong performance is that the powerful neural encoders are capable of capturing implicit output correlation of the tree structure~\cite{minimal-span, going-on-parser, two-local-model}.

\begin{figure}
    \centering
    \includegraphics[width=0.48\textwidth]{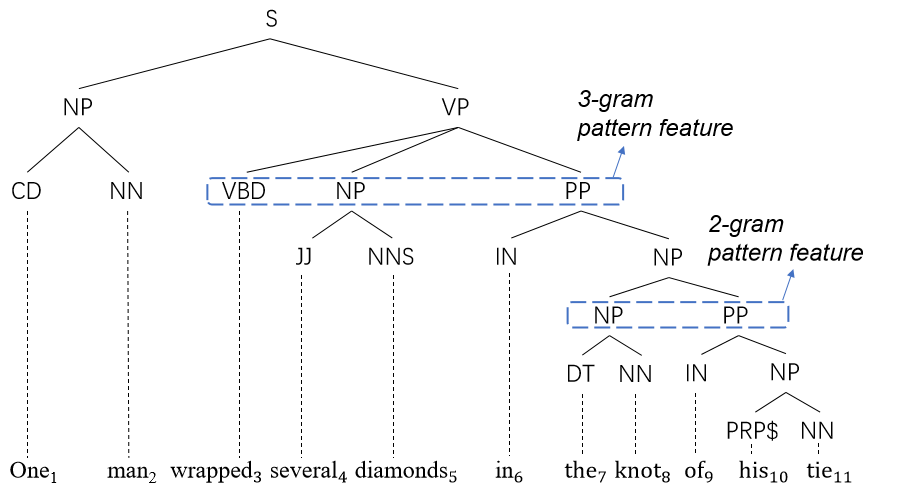}
    \caption{An example of the non-local $n$-gram {\it pattern} features: the 3-gram pattern $(3, 11, {\tt \{VBD \; NP \; PP\}})$ is composed of two constituent nodes and one part-of-speech node; the 2-gram pattern $(7, 11, {\tt \{NP \; PP\}})$ is composed of two constituent nodes. }
    \label{fig:intro}
\end{figure}

Recent work has shown that modeling non-local output dependencies can benefit neural structured prediction tasks, such as NER~\cite{lstm-crf}, CCG supertagging~\cite{lan} and dependency parsing~\cite{zhang-etal-2020-efficient}. 
Thus, an interesting research question is whether injecting non-local tree structure features is also beneficial to neural chart-based constituency parsing. 
To this end, we introduce two auxiliary training objectives. 
The first is {\it Pattern Prediction}. 
As shown in Figure~\ref{fig:intro}, we define {\it pattern} as the $n$-gram constituents sharing the same parent.\footnote{Patterns are mainly composed of $n$-gram constituents but also include part-of-speech tags as auxiliary.} We ask the model to predict the pattern based on its span representation, which directly injects the non-local constituent tree structure to the encoder. 

To allow stronger interaction between non-local patterns and local constituents, 
we further propose a {\it Consistency} loss, which regularizes the co-occurrence between constituents and patterns by collecting corpus-level statistics. In particular, we count whether the constituents can be a sub-tree of the pattern based on the training set.
For instance, both ${\tt NNS}$ and ${\tt NP}$ are legal to occur as sub-trees of the 3-gram pattern ${\{\tt VBD \ NP \ PP\}}$ in Figure~\ref{fig:intro}, while ${\tt S}$ or ${\tt ADJP}$ cannot be contained within this pattern based on grammar rules.
Similarly, for the 2-gram pattern ${\{\tt NP \ PP\}}$ highlighted in Figure~\ref{fig:intro}, both ${\tt IN}$ and ${\tt NP}$ are consistent constituents, but ${\tt JJ}$ is not.
The {\it Consistency} loss can be considered as injecting prior linguistic knowledge to our model, which forces the encoder to understand the grammar rules.  
Non-local dependencies among the constituents that share the same pattern are thus explicitly modeled. 
We denote our model as Injecting {\bf N}on-local {\bf F}eatures for neural {\bf C}hart-based parsers (NFC).

We conduct experiments on both PTB and CTB. Equipped with BERT, NFC achieves 95.92 F1 on PTB test set, which is the best reported performance for BERT-based single-model parsers. For Chinese constituency parsing, NFC achieves highly competitive results (92.31 F1) on CTB, outperforming the baseline self-attentive parser (91.98 F1) and a 0-th order neural CRF parser (92.27 F1)~\cite{constituency-crf}.
To further test the generalization ability, we annotate a multi-domain test set in English, including dialogue, forum, law, literature and review domains. Experiments demonstrate that NFC is robust in zero-shot cross-domain settings.
Finally, NFC also performs competitively with other languages using the SPMRL 2013/2014 shared tasks, establishing the best reported results on three rich resource languages. 
We release our code and models at \url{https://github.com/RingoS/nfc-parser}.





\section{Related Work}

\paragraph{Constituency Parsing.}
There are mainly two lines of approaches for constituency parsing. 
Transition-based methods process the input words sequentially and construct the output constituency tree incrementally by predicting a series of local transition actions \cite{zhang-clark-2009-transition, cross-huang-2016-span, liu-zhang-2017-order}. 
For these methods, the sequence of transition actions make traversal over a constituent tree.
Although transition-based methods directly model partial tree structures, their local decision nature may lead to error propagation \cite{error-propagation} and worse performance compared with methods that model long-term dependencies \cite{analyzing-dependency, zhang-nivre-2012-analyzing}. 
Similar to transition-based methods, NFC also directly models partial tree structures. 
The difference is that we inject tree structure information using two additional loss functions. 
Thus, our integration of non-local constituent features is implicit in the encoder, rather than explicit in the decoding process. 
While the relative effectiveness is empirical, it could potentially alleviate error propagation.

Chart-based methods score each span independently and perform global search over all possible trees to find the highest-score tree given a sentence. 
\citet{crf-parsing} represented nonlinear features to a traditional CRF parser computed with a feed-forward neural network. 
\citet{stern-etal-2017-minimal} first used LSTM to represent span features. 
\citet{san-constituency} adopted a self-attentive encoder instead of the LSTM encoder to boost parser performance.
\citet{label-attention-parsing} proposed label attention layers to replace self-attention layers.
\citet{head-driven} integrated constituency and dependency structures into head-driven phrase structure grammar.
\citet{span-attention} used span attention to produce span representation to replace the subtraction of the hidden states at the span boundaries.
Despite their success, above work mainly focuses on how to better encode features over the input sentence. In contrast, we take the encoder of \citet{san-constituency} intact, being the first to explore new ways to introduce non-local training signal into the local neural chart-based parsers.

\paragraph{Modeling Label Dependency.}
There is a line of work focusing on modeling non-local output dependencies. 
\citet{mll} used a Bayesian network to encode the label dependency in multi-label learning.
For neural sequence labeling, \citet{zhou-xu-2015-end} and \citet{lstm-crf} built a CRF layer on top of neural encoders to capture label transition patterns.
\citet{forest-ner} introduced a sentence-level constraint to encourage the model to generate coherent NER predictions. \citet{lan} investigated label attention network to model the label dependency by producing label distribution in sequence labeling tasks. \citet{gui-etal-2020-uncertainty} proposed a two-stage label decoding framework based on Bayesian network to model long-term label dependencies. For syntactic parsing, \citet{constituency-crf} demonstrated that structured Tree CRF can boost parsing performance over graph-based dependency parser. 
Our work is in line with these in the sense that we consider non-local structure information for neural structure prediction.
To our knowledge, we are the first to inject sub-tree structure into neural chart-based encoders for constituency parsing.

\section{Baseline}
\label{sec:baseline}
Our baseline is adopted from the parsing model of \citet{san-constituency} and \citet{kitaev-etal-2019-multi-lingual}. 
Given a sentence $X=\{x_1,...,x_n\}$, its corresponding constituency parse tree $T$ is composed by a set of labeled spans 
\begin{equation}
    T = \{(i_t,j_t,l^{\mathrm{c}}_t)\}|_{t=1}^{|T|}
\end{equation}
where $i_t$ and $j_t$ represent the $t$-{th} constituent span's fencepost positions and $l^{\mathrm{c}}_t$ represents the constituent label. 
The model assigns a score $s(T)$ to tree $T$, which can be decomposed as
\begin{equation}
   s(T) = \sum_{(i,j,l)\in T} s(i,j,l^{\mathrm{c}})
\end{equation}

Following~\citet{kitaev-etal-2019-multi-lingual}, we use BERT with a self-attentive encoder as the scoring function $s(i, j, \cdot)$, and a chart decoder to perform a global-optimal search over all possible trees to find the highest-scoring tree given the sentence.
In particular, given an input sentence $X=\{x_1,...,x_n\}$, a list of hidden representations $\mathbf{H}_1^n = \{\mathbf{h}_1,\mathbf{h}_2,\dots,\mathbf{h}_n\}$ is produced by the encoder, where $\mathbf{h}_i$ is a hidden representation of the input token $x_i$. 
Following previous work, the representation of a span $(i,j)$ is constructed by:
\begin{equation}
    \mathbf{v}_{i,j} = \mathbf{h}_j - \mathbf{h}_i
\label{eq:span_rep}
\end{equation}

Finally, $\mathbf{v}_{i,j}$ is fed into an MLP to produce real-valued scores $s(i,j,\cdot)$ for all constituency labels: 
\begin{equation}
\fontsize{10}{12}\selectfont
    s(i,j,\cdot) = \mathbf{W}^{\mathrm{c}}_2  \textsc{ReLU}(\mathbf{W}^{\mathrm{c}}_1 \mathbf{v}_{i,j} + \mathbf{b}^{\mathrm{c}}_1) + \mathbf{b}^{\mathrm{c}}_2
\label{eq:mlp_1}
\end{equation}
where 
$\mathbf{W}^{\mathrm{c}}_1$, $\mathbf{W}^{\mathrm{c}}_2$, $\mathbf{b}^{\mathrm{c}}_1$ and $\mathbf{b}^{\mathrm{c}}_2$ are trainable parameters, 
$\mathbf{W}^{\mathrm{c}}_2 \in \mathbb{R}^{|H| \times |L^\mathrm{c}|}$ can be considered as the constituency label embedding matrix~\cite{lan}, where each column in $\mathbf{W}_2^c$ corresponds to the embedding of a particular constituent label. $|H|$ represents the hidden dimension and $|L^\mathrm{c}|$ is the size of the constituency label set. 

\paragraph{Training.}
The model is trained to satisfy the margin-based constraints 
\begin{equation}
    s(T^*) \geq s(T) + \Delta(T, T^*)
\end{equation}
where $T^*$ denotes the gold parse tree, and $\Delta$ is Hamming loss. 
The hinge loss can be written as
\begin{equation}
\small
    \mathcal{L}_{ \mathrm{cons}} = \max \big(0, \max_{T \neq T^*}[s(T) + \Delta(T, T^*)] - s(T^*)\big)
    \label{eq:margin_loss}
\end{equation}

During inference time, the most-optimal tree
\begin{equation}
    \hat{T} = \argmax_T s(T)
\end{equation}
is obtained using a CKY-like algorithm. 

\section{Additional Training Objectives}
We propose two auxiliary training objectives to inject non-local features into the encoder, which rely only on the annotations in the constituency treebank, but not external resources. 

\begin{figure}
    \centering
    \includegraphics[width=0.5\textwidth]{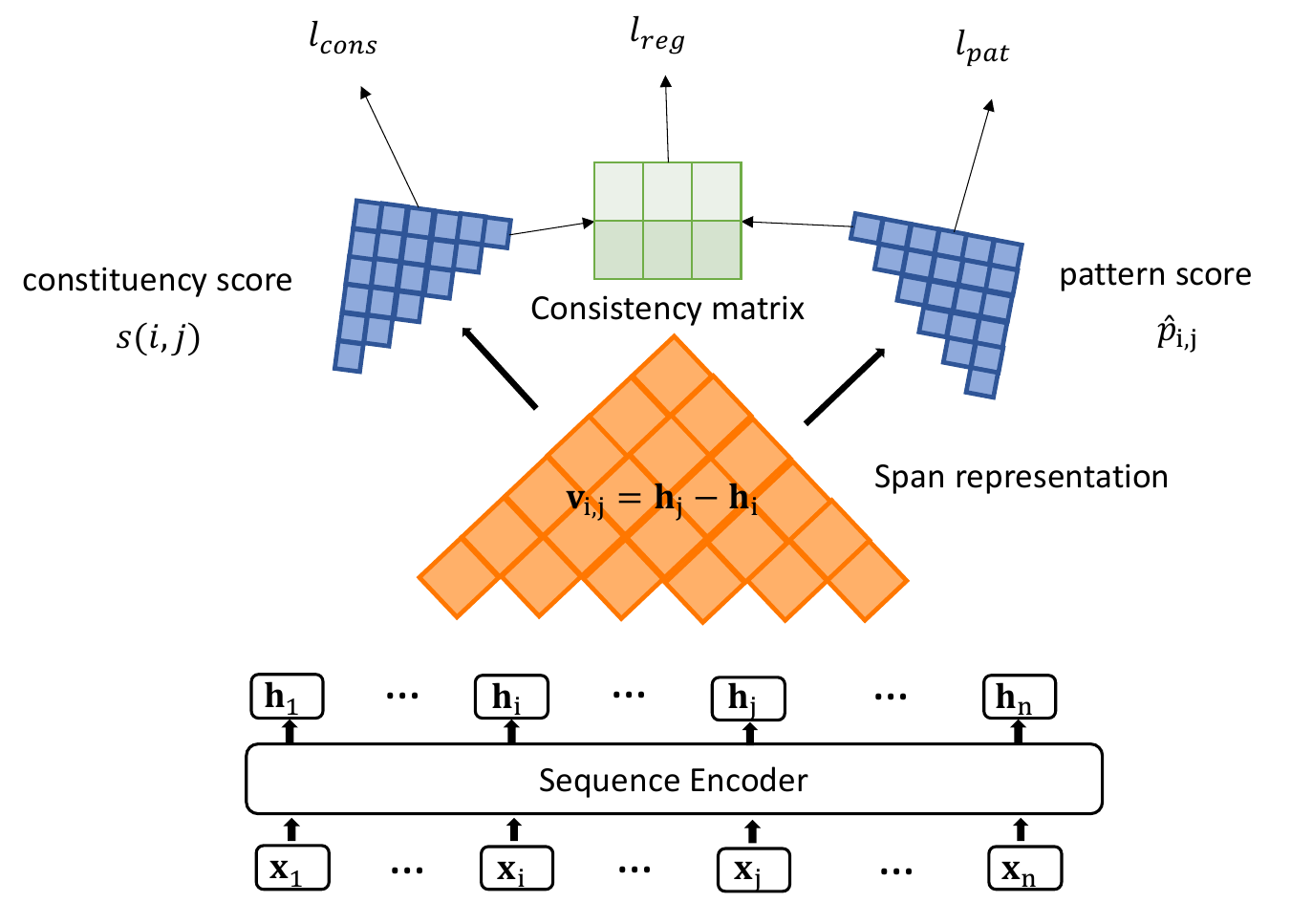}
    \caption{The three training objectives in NFC.}
    \label{fig:method}
\end{figure}

\subsection{Instance-level Pattern Loss}
\label{sec:pattern_loss}
We define $n$-gram constituents, which shares the same parent node, as a pattern. 
We use a triplet $(i^\mathrm{p}, j^\mathrm{p}, l^\mathrm{p})$ to denote a pattern span beginning from the $i^\mathrm{p}$-th word and ending at $j^\mathrm{p}$-th word. 
$l^\mathrm{p}$ is the corresponding pattern label. 
Given a constituency parse tree in Figure~\ref{fig:intro}, $(3, 11, {\tt \{VBD \; NP \; PP\}})$ is a $3$-gram pattern.

Similar to Eq~\ref{eq:mlp_1}, an MLP is used for transforming span representations to pattern prediction probabilities: 
\begin{equation}
\small
    \hat{p}_{i,j} = \softmax\big(\mathbf{W}^{\mathrm{p}}_2  \textsc{ReLU}(\mathbf{W}^{\mathrm{p}}_1 \mathbf{v}_{i,j} + \mathbf{b}^{\mathrm{p}}_1) + \mathbf{b}^{\mathrm{p}}_2\big)
\label{softmax}
\end{equation}
where $\mathbf{W}^{\mathrm{p}}_1$, $\mathbf{W}^{\mathrm{p}}_2$, $\mathbf{b}^{\mathrm{p}}_1$ and $\mathbf{b}^{\mathrm{p}}_2$ are trainable parameters,  $\mathbf{W}^{\mathrm{p}}_2 \in \mathbb{R}^{|H| \times |L^{\mathrm{p}}|}$ can be considered as the pattern label embedding matrix, where each column in $\mathbf{W}_2^p$ corresponds to the embedding of a particular pattern label. $|L^{\mathrm{p}}|$ represents the size of the pattern label set.
For each instance, the cross-entropy loss between the predicted patterns and the gold patterns are calculated as
\begin{equation}
   \mathcal{L}_{pat} = - \sum_{i=1}^n \sum_{j=1}^n p_{i,j} \log \hat{p}_{i,j}
\label{eq:pat}
\end{equation}

We use the span-level cross-entropy loss for patterns (Eq~\ref{eq:pat}) instead of the margin loss in Eq~\ref{eq:margin_loss}, because our pattern-prediction objective aims to augment span representations via greedily classifying each pattern span, rather than to reconstruct the constituency parse tree through dynamic programming. 

\subsection{Corpus-level Consistency Loss}
\label{sec:consistency_loss}
Constituency scores and pattern probabilities are produced based on a shared span representation; however, the two are subsequently separately predicted. 
Therefore, although the span representations contain both constituent and pattern information, the dependencies between constituent and pattern predictions are not explicitly modeled. 
Intuitively, constituents are distributed non-uniformly in patterns, and such correlation can be obtained in the corpus-level statistic. We propose a consistency loss, which explicitly models the non-local dependencies among constituents that belong to the same pattern. As mentioned in the introduction, we regard all constituent spans within a pattern span as being consistent with the pattern span. Take 2-gram patterns for example, which represents two neighboring subtrees covering a text span. The constituents that belong to the two subtrees, including the top constituent and internal sub constituents, are considered as being consistent. We consider only the constituent labels but not their corresponding span locations for this task.

This loss can be understood first at the instance level.
In particular, if a constituent span $(i_t,j_t,l^{\mathrm{c}}_t)$ is a subtree of a pattern span $(i_{t'}, j_{t'}, l^{\mathrm{p}}_{t'})$, i.e. $i_t >= i_{t'} $ and $j_t <= j_{t'}$, 
where $l^{\mathrm{c}}_t=L^{\mathrm{c}}[a]$ (the $a$-th constituent label in $L^{\mathrm{c}}$) and $l^{\mathrm{p}}_{t'}=L^{\mathrm{p}}[b]$ (the $b$-th pattern label in $L^p$),
we define $L^{\mathrm{c}}[a]$ and $L^{\mathrm{p}}[b]$ to be {\it consistent} (denoted as $y_{a,b}=1$). 
Otherwise we consider it to be {\it non-consistent} (denoted as $y_{a,b}=0$). 
This yields a consistency matrix $\mathbf{Y} \in \mathbb{R}^{|L^\mathrm{c}| \times |L^{\mathrm{p}}|}$ for each instance. 
The gold consistency matrix $\mathbf{Y}$ provides information regarding non-local dependencies among constituents and patterns. 




An intuitive method to predict the consistency matrix $\mathbf{Y}$ is to make use of the constituency label embedding matrix ${\mathbf{W}^{\mathrm{p}}_2}$ (see Eq~\ref{eq:mlp_1} for definition), the pattern label embedding matrix $\mathbf{W}^{\mathrm{c}}_2$ (see Eq~\ref{softmax} for definition) and the span representations $\mathbf{V}$ (see Eq~\ref{eq:span_rep} for definition): 
\begin{equation}
    \hat{\mathbf{Y}} = \mathrm{Sigmoid}\big(({\mathbf{W}^{\mathrm{c}}_2}^\mathsf{T} \mathbf{U}_1 \mathbf{V}) (\mathbf{V}^\mathsf{T} \mathbf{U}_2 {\mathbf{W}^{\mathrm{p}}_2})\big)
\label{eq:constraint}
\end{equation}
where $\mathbf{U}_1, \mathbf{U}_2 \in \mathbb{R}^{|H| \times |H|}$ are trainable parameters.

Intuitively, the left term, ${\mathbf{W}^{\mathrm{c}}_2}^\mathsf{T} \mathbf{U}_1 \mathbf{V}$, integrates the representations of the pattern span and all possible constituent label embeddings. The second term, $\mathbf{V}^\mathsf{T} \mathbf{U}_2 {\mathbf{W}^{\mathrm{p}}_2}$, integrates features of the span and all pattern embeddings. Each binary element in the resulting $\hat{\mathbf{Y}} \in \mathbb{R}^{|L^c|\times|L^p|}$ denotes whether a particular constituent label is consistent with a particular pattern in the given span context.
Eq~\ref{eq:constraint} can be predicted on the instance-level for ensuring consistency between patterns and constituent.
However, this naive method is difficult for training, and computationally infeasible, because the span representation matrix $\mathbf{V} \in \mathbb{R}^{|H| \times n^2}$ is composed of $n^2$ span representations $\mathbf{v}_{i,j} \in \mathbb{R}^{|H|}$ and the asymptotic complexity is:
\begin{equation}
    O\Big((|L^{\mathrm{p}}| + |L^\mathrm{c}|) (|H|^2 + n^2 |H|) + |L^{\mathrm{p}}||L^\mathrm{c}|n^2\Big)
\end{equation}
for a single training instance. 

We instead use a corpus-level constraint on the non-local dependencies among constituents and patterns.
In this way, Eq~\ref{eq:constraint} is reduced to be independent of individual span representations:
\begin{equation}
    \hat{\mathbf{Y}} = \mathrm{Sigmoid}\big(\mathbf{W}^{\mathrm{c}}_2 \mathbf{U} {\mathbf{W}^{\mathrm{p}}_2}^\mathsf{T}\big)
\label{eq:consis}
\end{equation}
where $\mathbf{U} \in \mathbb{R}^{|H| \times |H|}$ is trainable.

This trick decreases the asymptotic complexity to 
$O(|L^{\mathrm{c}}||H|^2 + |L^\mathrm{p}||L^\mathrm{c}||H|)$.
The cross-entropy loss between the predicted consistency matrix and gold consistency labels is used to optimize the model:
\begin{equation}
    \mathcal{L}_{reg} = - \sum_{a=1}^{|L^\mathrm{c}|} \sum_{b=1}^{|L^{\mathrm{p}}|} y_{a,b} \log \hat{y}_{a,b}
\end{equation}

The corpus-level constraint can be considered as a prior linguistic knowledge statistic from the treebank, which forces the encoder to understand the grammar rules.



\subsection{Training}
Given a constituency tree, we minimize the sum of the three objectives to optimize the parser: 
\begin{equation}
    \mathcal{L} = \mathcal{L}_{cons} + \mathcal{L}_{pat} + \mathcal{L}_{reg}
\end{equation}

\subsection{Computational Cost}
\label{sec:cost}
The number of training parameters increased by NFC is $\mathbf{W}^{\mathrm{p}}_1 \in \mathbb{R}^{|H| \times |H|}$, $\mathbf{W}^{\mathrm{p}}_2 \in \mathbb{R}^{|H| \times |L^p|}$ , $\mathbf{b}^{\mathrm{p}}_1 \in \mathbb{R}^{|H|}$ and $\mathbf{b}^{\mathrm{p}}_2 \in \mathbb{R}^{|L^p|}$ in Eq~\ref{softmax} and $\mathbf{U} \in \mathbb{R}^{|H| \times |H|}$ in Eq~\ref{eq:consis}. Taking training model on PTB as an example, NFC adds less than 0.7M parameters to 342M parameters baseline model \cite{san-constituency} based on BERT-large-uncased during training.
NFC is identical to our baseline self-attentive parser \cite{san-constituency} during inference.


\section{Experiments}
\label{sec:exp}

We empirically compare NFC with the baseline parser in different settings, including in-domain, cross-domain and multilingual benchmarks.



\begin{table}[t!]
    \centering
    \begin{adjustbox}{max width=1.0\columnwidth}
    \begin{tabular}{ccccc}
    \hline
    Data & Lang / Domain & \# Train & \# Dev & \# Test \\
    \hline
    PTB & English & 39,832 & 1,700 & 2,416 \\
    CTB & Chinese & 17,544 & 352 & 348 \\
    SPMRL & French & 14,759 & 1,235 & 2,541 \\
    SPMRL & German & 40,472 & 5,000 & 5,000 \\
    SPMRL & Korean & 23,010 & 2,066 & 2,287 \\
    SPMRL & Basque & 7,577 & 948 & 946 \\
    SPMRL & Polish & 6,578 & 821 & 822 \\
    SPMRL & Hungarian & 8,146 & 1,051 & 1,009 \\
    \hline
    MCTB & Dialogue & - & - & 1,000 \\
    MCTB & Forum & - & - & 1,000 \\
    MCTB & Law & - & - & 1,000 \\
    MCTB & Literature & - & - & 1,000 \\
    MCTB & Review & - & - & 1,000 \\
    \hline
    \end{tabular}
    \end{adjustbox}
    \caption{Dataset statistics. \# - number of sentences.}
    \label{tab:data_statistics}
\end{table}

\subsection{Dataset}
\label{sec:dataset}
Table~\ref{tab:data_statistics} shows the detailed statistic of our datasets.

\paragraph{In-domain.} We conduct experiments on both English and Chinese, using the Penn Treebank~\cite{ptb} as our English dataset, with standard splits of section 02-21 for training, section 22 for development and section 23 for testing. For Chinese, we split the Penn Chinese Treebank (CTB) 5.1~\cite{ctb}, taking articles 001-270 and 440-1151 as training set, articles 301-325 as development set and articles 271-300 as test set. 


\paragraph{Cross-domain.}
To test the robustness of our methods across difference domains, we further annotate five test set in dialogue, forum, law, literature and review domains. 
For the dialogue domain, we randomly sample dialogue utterances from Wizard of Wikipedia \cite{wizard}, which is a chit-chat dialogue benchmark produced by humans. 
For the forum domain, we use users' communication records from Reddit, crawled and released by \citet{voelske:2017}.
For the law domain, we sample text from European Court of Human Rights Database \cite{ECtHR}, which includes detailing judicial decision patterns. 
For the literature domain, we download literary fictions from Project Gutenberg\footnote{\url{https://www.gutenberg.org/}}. 
For the review domain, we use plain text across a variety of product genres, released by SNAP Amazon Review Dataset \cite{review}. After obtaining the plain text, we ask annotators whose majors are linguistics to annotate constituency parse tree by following the PTB guideline. 
We name our dataset as {\bf M}ulti-domain {\bf C}onstituency {\bf T}ree{\bf b}ank (MCTB).
More details of the dataset are documented in \citet{yang-etal-2022-challenge}.

\paragraph{Multi-lingual.} For the multilingual testing, we select three rich resource language from the SPMRL 2013-2014 shared task \cite{spmrl}: French, German and Korean, which include at least 10,000 training instances, and three low-resource language: Hungarian, Basque and Polish.





\subsection{Setup}
Our code is based on the open-sourced code of~\citet{san-constituency}\footnote{Available at~\url{https://github.com/nikitakit/self-attentive-parser}.}. 
The training process gets terminated if no improvement on development F1 is obtained in the last 60 epochs. 
We evaluate the models which have the best F1 on the development set. 
For fair comparison, all reported results and baselines are augmented with BERT. We adopt {\tt BERT-large-uncased} for English, {\tt BERT-base} for Chinese and {\tt BERT-multi-lingual-uncased} for other languages.
Most of our hyper-parameters are adopted from \citet{san-constituency} and \citet{cross-domain-parser}. 
For scales of the two additional losses, we set the scale of pattern loss to 1.0 and the scale of consistency loss to 5.0 for all experiments. 

To reduce the model size, we filter out those non-local pattern features that appear less than 5 times in the PTB training set and  those that account for less than 0.5\% of all pattern occurrences in the CTB training set. The out-of-vocabulary patterns are set as ${\tt <UNK>}$.
This results in moderate pattern vocabulary sizes of 841 for PTB and 514 for CTB. 
For evaluation on PTB, CTB and cross-domain dataset, we use the EVALB script for evaluation.
For the SPMRL datasets, we follow the same setup in EVALB as \citet{san-constituency}.

\subsection{In-domain Experiments}

\begin{table}[!t]
    \centering
    \begin{adjustbox}{max width=1.0\columnwidth}
    \begin{tabular}{lccc}
        \hline
        {\bf Model} & {\bf LR} & {\bf LP} & {\bf F1} \\
        \hline
        \citet{liu-zhang-2017-order} $\diamond$ & - & - & 95.71 \\
        \citet{san-constituency} & 95.46 & 95.73 & 95.59 \\
        \citet{head-driven} & 95.51 & 95.93 & 95.72 \\
        \citet{head-driven} * & 95.70 & 95.98 & 95.84 \\
        \citet{constituency-crf} & 95.53 & 95.85 & 95.69 \\
        \citet{constituency-point} & - & - & 95.48 \\
        \citet{span-attention} & 95.58 & 96.11 & 95.85 \\
        \hline
        \multicolumn{4}{c}{This work} \\
        \hline
        \citet{san-constituency} $\dagger$ & 95.56 & 95.89 & 95.72 \\
        NFC w/o $\mathcal{L}_{reg}$ & 95.49 & 96.07 & 95.78 \\
        NFC & {\bf 95.70} & {\bf 96.14} & {\bf 95.92} \\
        \hline
    \end{tabular}
    \end{adjustbox}
    \caption{Performance (w/ BERT) on the test set of PTB. $\dagger$ indicates our reproduced results, which is also the baseline that our method is built upon. * indicates training with extra supervision from dependency parsing data. $\diamond$ indicates that the results are reported by the re-implementation of \citet{cross-domain-parser}. }
    \label{tab:ptb}
\end{table}

\begin{table}[t!]
    \centering
    \begin{adjustbox}{max width=1.0\columnwidth}
    \begin{tabular}{lccc}
    \hline
        {\bf Model} & {\bf LR} & {\bf LP} & {\bf F1} \\
        \hline
        \citet{liu-zhang-2017-order} $\diamond$ & - & - & 91.81 \\
        \citet{san-constituency} & 91.55 & 91.96 & 91.75 \\
        \citet{constituency-crf} & 92.04 & 92.51 & 92.27 \\
        \citet{head-driven}  & 91.14 & 93.09 & 92.10 \\
        \citet{span-attention} & 92.14 & 92.25 & 92.20 \\
        \hline
        \multicolumn{4}{c}{This work} \\
        \hline
        \citet{san-constituency} $\dagger$ & 91.80 & 92.23 & 91.98 \\
        NFC w/o $\mathcal{L}_{reg}$ & 91.87 & 92.40 & 92.13 \\
        NFC & {\bf 92.17} & {\bf 92.45} & {\bf 92.31} \\
        \hline
        \multicolumn{4}{c}{w/ External Dependency Supervision} \\
        \hline
        \citet{head-driven} * & 92.03 & 92.33 & 92.18 \\
        \citet{label-attention-parsing}* & 91.85 & 93.45 & 92.64 \\
        
        \hline
    \end{tabular}
    \end{adjustbox}
    \caption{Constituency parsing performance (w/ BERT) on the test set of CTB 5.1. The symbols ($\dagger$, * and $\diamond$) are explained in Table~\ref{tab:ptb}.}
    \label{tab:ctb}
\end{table}

\begin{table*}[th!]
    \fontsize{10}{12}\selectfont
    \centering
    \begin{tabular}{c|c|cccccc|c}
        \hline
        \multirow{2}{*}{\bf Model} & {\bf In-domain} &  & \multicolumn{5}{c}{\bf Cross-domain} \\
        \cline{2-9}
         & \bf PTB & \bf Bio & \bf Dialogue & \bf Forum &\bf Law & \bf Literature & \bf Review & \bf Avg \\
        \hline
        \citet{liu-zhang-2017-order}     & 95.65 & 86.33 & 85.56 & 85.42 & 91.50 & 84.84 & 83.53 & 86.20\\
        \citet{san-constituency}         & 95.72 & 86.61 & 86.30 & 86.29 & 92.08 & 86.10 & 83.88 & 86.88 \\
    \hline
        NFC & {\bf 95.92} & 86.43 & {\bf 89.85} & {\bf 88.52} & {\bf 95.43} & {\bf 90.75} & {\bf 88.10} & {\bf 89.85} \\
        \hline
    \end{tabular}
    \caption{Constituency parsing results with BERT (F1 scores) on the cross-domain test set. }
    \label{tab:cross-domain}
\end{table*}

\begin{table*}[t!]
    \begin{center}
            \small
    \begin{tabular}{c|ccc|c|ccc|c|c}
        \hline
        \multirow{2}{*}{\bf Model} & \multicolumn{4}{c|}{\bf Rich resource} & \multicolumn{4}{c|}{\bf Low Resource} & \multirow{2}{*}{\bf Avg} \\
       \cline{2-9}
        & \bf French & \bf German & \bf Korean & \bf Avg & \bf Hungarian & \bf Basque & \bf Polish & \bf Avg & \\
        \hline
        \citet{san-constituency} & 87.42 & 90.20 & 88.80 & 88.81 & 94.90 & 91.63 & \bf 96.36 & 94.30 & 91.55 \\
         \citet{constituency-point} & 86.69 & 90.28 & 88.71 & 88.56 & 94.24  & \bf 92.02 & 96.14 & 94.13 & 91.34 \\
         \hline
        \citet{san-constituency} $\dagger$ & 87.38 & 90.25 & 88.91 & 88.85 & 94.56 & 91.66 & 96.14 & 94.12 & 91.48 \\
         NFC & \bf 87.51 & \bf 90.43 & \bf 89.07 & \bf 89.00 & \bf 94.95 & 91.73 & 96.33 & {\bf 94.34} &{\bf 91.67} \\
        \hline
    \end{tabular}
    \end{center}
    \caption{Multilingual Experiment results on SPMRL test-sets. $\dagger$ indicates our reproduced baselines.}
    \label{tab:multi-lingual}
\end{table*}

We report the performance of our method on the test sets of PTB and CTB in Table~\ref{tab:ptb} and~\ref{tab:ctb}, respectively. 
Compared with the baseline parser~\cite{san-constituency}, our method obtains an absolute improvement of 0.20\% F1 on PTB ($p$\textless0.01) and 0.33\% F1 on CTB ($p$\textless0.01), 
which verifies the effectiveness of injecting non-local features into neural local span-based constituency parsers. 
Note that the proposed method adds less than 0.7M parameters to the 342M parameter baseline model using \mbox{BERT-large}.

The parser trained with both the pattern loss (Section~\ref{sec:pattern_loss}) and consistency loss (Section~\ref{sec:consistency_loss}) outperforms the one trained only with pattern loss by 0.14\% F1 ($p$\textless0.01). 
This suggests that the constraints between constituents and non-local pattern features are crucial for injecting non-local features into local span-based parsers. 
One possible explanation for the improvement is that the constraints may bridge the gap between local and non-local supervision signals, since these two are originally separately predicted while merely sharing the same encoder in the training phase.

We further compare our method with the recent state-of-the-art parsers on PTB and CTB. 
\citet{liu-zhang-2017-order} propose an in-order transition-based constituency parser. 
\citet{san-constituency} use self-attentive layers instead of LSTM layers to boost performance. 
\citet{head-driven} jointly optimize constituency parsing and dependency parsing objectives using head-driven phrase structure grammar. 
\citet{label-attention-parsing} extend \citet{head-driven} by introducing label attention layers.
\citet{constituency-crf} integrate a CRF layer to a chart-based parser for structural training (without non-local features). 
\citet{span-attention} use span attention for better span representation.


Compared with these methods, the proposed method achieves an F1 of 95.92\%, which exceeds previous best numbers for BERT-based single-model parsers on the PTB test set. We further compare  experiments for five runs, and find that NFC significantly outperforms \citet{san-constituency} ($p$\textless0.01).
The test score of 92.31\% F1 on CTB significantly outperforms the result (91.98\% F1) of the baseline ($p$\textless0.01).
Compared with the CRF parser of ~\citet{constituency-crf}, our method gives better scores without global normalization in training. 
This shows the effectiveness of integrating non-local information during training using our simple regularization. 
The result is highly competitive with the current best result~\cite{label-attention-parsing}, which is obtained by using external dependency parsing data.

\subsection{Cross-domain Experiments}

We compare the generalization of our methods with baselines in Table~\ref{tab:cross-domain}. 
In particular, all the parsers are trained on PTB training and validated on PTB development, and are tested on cross-domain test in the zero-shot setting. 
As shown in the table, our model achieves 5 best-reported results among 6 cross-domain test sets with an averaged F1 score of 89.85\%, outperforming our baseline parser by 2.97\% points. 
This shows that structure information is useful for improving cross-domain performance, which is consistent with findings from previous work~\cite{cross-domain-parser}.

\begin{figure}[t!]
    \centering
    \includegraphics[width=0.5\textwidth]{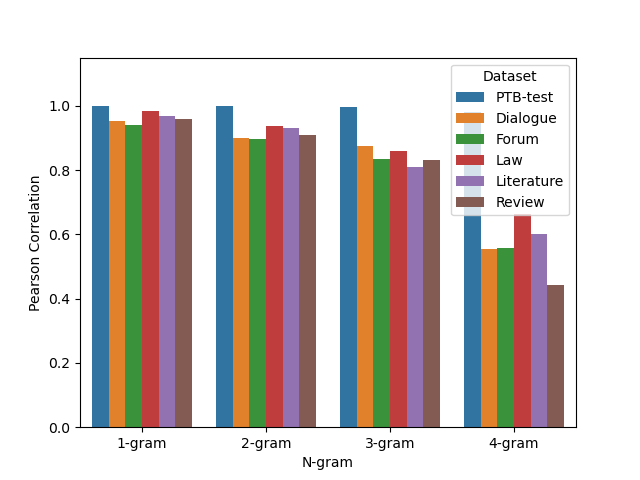}
    \caption{Pearson correlation of $n$-gram pattern distribution between PTB training set and different test set.}
    \label{fig:correlation}
\end{figure}

To better understand the benefit of pattern features, we calculate Pearson correlation of $n$-gram pattern distributions between the PTB training set and various test sets in Figure~\ref{fig:correlation}.
First, we find that the correlation between the PTB training set and the PTB test set is close to 1.0, which verifies the effectiveness of the corpus-level pattern knowledge during inference. 
Second, the $3$-gram pattern correlation of all domains exceeds 0.75, demonstrating that $n$-gram pattern knowledge is robust across domains, which supports the strong performance of NFC in the zero-shot cross-domain setting. 
Third, pattern correlation decreases significantly as $n$ increases, which suggests that transferable non-local information is limited to a certain window size of $n$-gram constituents. 

\begin{figure}[t!]
    \centering
    \subfigure[F1 scores measured by 3-gram pattern.]{
    \includegraphics[width=0.48\textwidth]{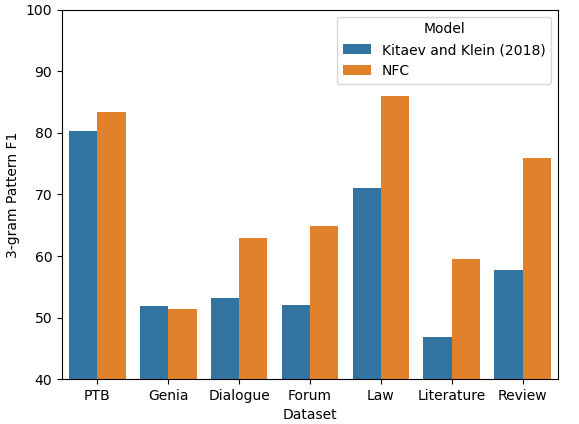}
    }
    \subfigure[F1 scores measured by 2-gram pattern.]{
    \includegraphics[width=0.48\textwidth]{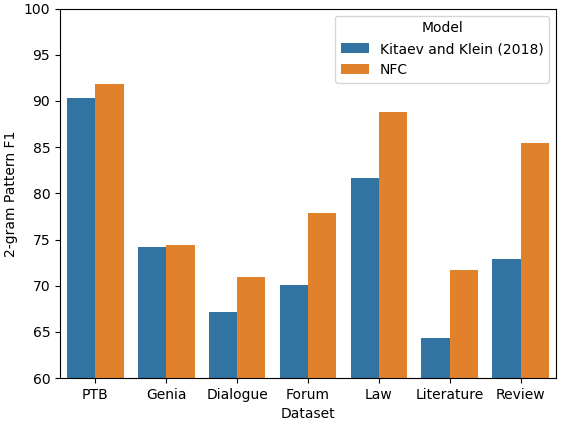} 
    }
    \caption{Pattern-level F1 on different English datasets. Noted that we train NFC based on 3-gram pattern in English. There is no direct supervision signal for 2-gram pattern.}
    \label{fig:pattern}
\end{figure}


\subsection{Multilingual Experiments}
We compare NFC with \citet{san-constituency} and \citet{constituency-point} on SPMRL. 
The results are shown in Table~\ref{tab:multi-lingual}. 
\citet{constituency-point} use pointer network to predict a sequence of pointing decisions for constituency parsing. As can be seen, \citet{constituency-point} do not show obvious advantages over \citet{san-constituency}.
NFC outperforms these two methods on three rich resource languages. 
For example, NFC achieves 89.07\% F1 on Korean, outperforming \citet{san-constituency} by 0.27\% F1, suggesting that NFC is generally effective across languages. However, NFC does not give better results compared with \citet{san-constituency} on low-resource languages. 
One possible explanation is that it is difficult to obtain prior linguistic knowledge from corpus-level statistics by using a relatively small number of instances.



\section{Analysis}

\subsection{$n$-gram Pattern Level Performance}

Figure~\ref{fig:pattern} shows the pattern-level F1 before and after introducing the two auxiliary training objectives. In particular, we calculate the pattern-level F1 by calculating the F1 score for patterns based on the constituency trees predicted by CKY decoding. Although our baseline parser with BERT achieves 95.76\% F1 scores on PTB, the pattern-level F1 is 80.28\% measured by 3-gram. When testing on the dialogue domain, the result is reduced to only 57.47\% F1, which indicates that even a strong neural encoder still has difficulties capturing constituent dependency from the input sequence alone. 
After introducing the pattern and consistency losses, NFC significantly outperforms the baseline parser measured by 3-gram pattern F1. 
Though there is no direct supervision signal for 2-gram pattern, NFC also gives better results on pattern F1 of 2-gram, which are subsumed by 3-gram patterns.
This suggests that NFC can effectively represent sub-tree structures.

\subsection{F1 against Span Length}

\begin{figure}[t!]
    \centering
    \includegraphics[width=0.48\textwidth]{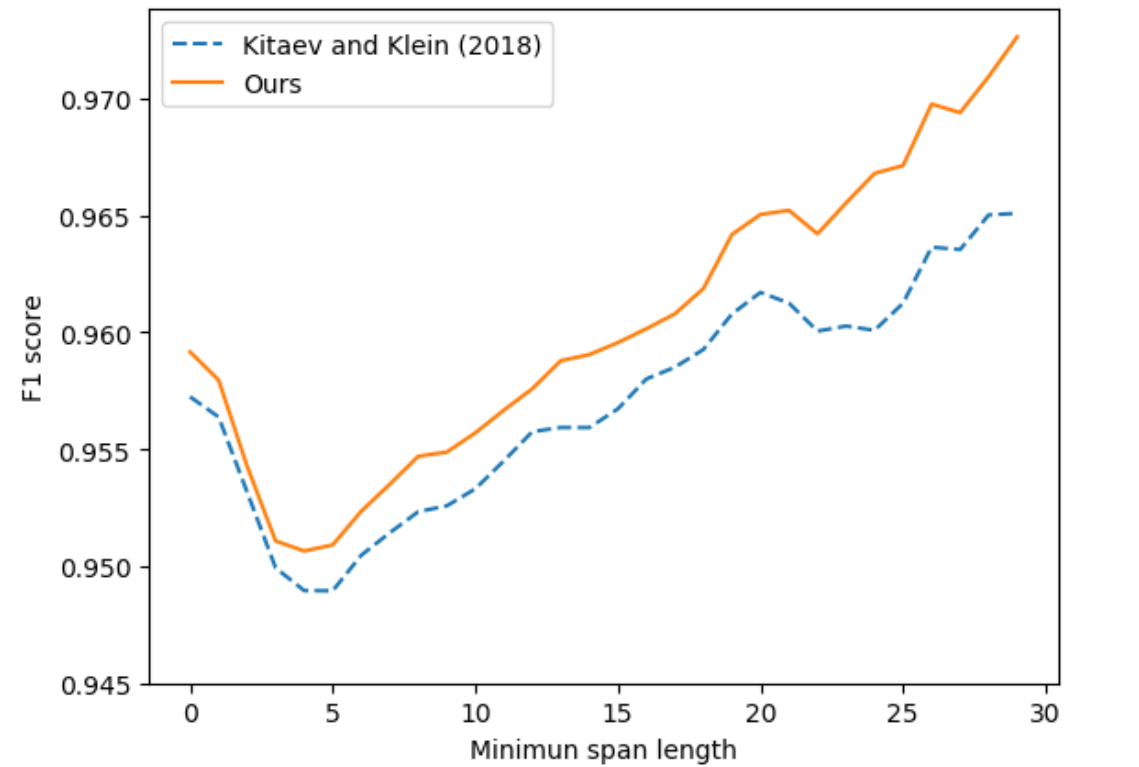}
    \caption{F1 scores versus minimum constituent span length on PTB test set. Note that constituent spans shorter than 30 accounts for approximately 98.5\% of all for the PTB test set. }
    \label{fig:span_length}
    \vspace{-0.3cm}
\end{figure}


We compare the performance of the baseline and our method on constituent spans with different word lengths. 
Figure~\ref{fig:span_length} shows the trends of F1 scores on the PTB test set as the minimum constituent span length increases. 
Our method shows a minor improvement at the beginning, but the gap becomes more evident when the minimum span length increases, demonstrating its advantage in capturing
more sophisticated constituency label dependency.

\begin{figure}[t!]
    \centering
    \includegraphics[width=0.48\textwidth]{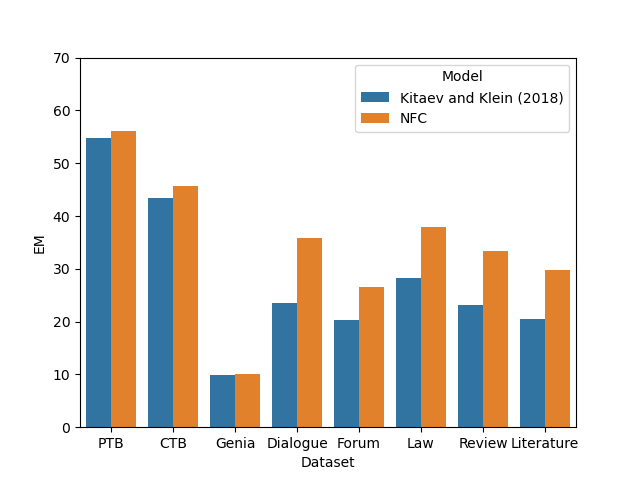}
    \caption{Exact matching (EM) score across different domains. EM indicates the percentage of sentences whose predicted trees are entirely correct.}
    \label{fig:em}
    \vspace{-0.4cm}
\end{figure}

\subsection{Exact Match}

Exact match score represents the percentage of sentences whose predicted trees are entirely the same as the golden trees. 
Producing exactly matched trees could improve user experiences in practical scenarios and benefit downstream applications on other tasks~\cite{em,kummerfeld-etal-2012-parser}.
We compare exact match scores of NFC with that of the baseline parser.
As shown in Figure~\ref{fig:em}, NFC achieves large improvements in exact match score for all domains.
For instance, NFC gets 33.40\% exact match score in the review domain, outperforming the baseline by 10.2\% points.
We assume that this results from the fact that NFC successfully ensures the output tree structure by modeling non-local correlation.

\subsection{Model Efficiency}
As mentioned in Section~\ref{sec:cost}, NFC only introduces a few training parameters to the baseline model~\cite{san-constituency}. 
For PTB, NFC takes about 19 hours to train with a single RTX 2080Ti, while the baseline takes about 13 hours. 
For CTB, the approximate training time is 12 hours for NFC and 7 hours for the baseline.
Our inference time is the same as that of the baseline parser, since no further computational operations are added to the inference phase. 
Both take around 11 seconds to parse the PTB section 23 (2416 sentences, an average of 23.5 tokens per sentence).

\section{Conclusion}

We investigated graph-based constituency parsing with non-local features -- both in the sense that features are not restricted to one constituent, and in the sense that they are not restricted to each training instance.
Experimental results verify the effectiveness of injecting non-local features to neural chart-based constituency parsing. 
Equipped with pre-trained BERT, our method achieves 95.92\% F1 on PTB and 92.31\% F1 on CTB. We further demonstrated that the proposed method gives better or competitive results in multilingual and zero-shot cross-domain settings.

\section*{Acknowledgements}
We appreciate the insightful comments from the anonymous reviewers.
We thank Zhiyang Teng for the insightful discussions. 
We gratefully acknowledge funding from the National Natural Science Foundation of China (NSFC No.61976180).

\section*{Ethical Considerations}

As mentioned in Section~\ref{sec:dataset}, we collected the raw data from free and publicly available sources that have no copyright or privacy issues.
We recruited our annotators from the linguistics departments of local universities through public advertisement with a specified pay rate. 
All of our annotators are senior undergraduate students or graduate students in linguistic majors who took this annotation as a part-time job. 
We manually shuffled the data so that all batches of to-be-annotated data have similar lengths on average.
An annotator could annotate around 25 instances per hour. We pay them 50 CNY an hour.
The local minimum salary in the year 2021 is 22 CNY per hour for part-time jobs.

Our annotated data only involves factual information (i.e., syntactic annotation), but not opinions, attitudes or beliefs. 
Therefore, the annotation job does not belong to human subject research; and IRB approval is not required.



\bibliographystyle{acl_natbib}
\bibliography{custom}

\begin{thebibliography}{39}
\expandafter\ifx\csname natexlab\endcsname\relax\def\natexlab#1{#1}\fi

\bibitem[{Chen et~al.(2017)Chen, Zhu, Ling, Wei, Jiang, and Inkpen}]{nli}
Qian Chen, Xiaodan Zhu, Zhen-Hua Ling, Si~Wei, Hui Jiang, and Diana Inkpen.
  2017.
\newblock \href {https://doi.org/10.18653/v1/P17-1152} {Enhanced {LSTM} for
  natural language inference}.
\newblock In \emph{Proceedings of the 55th Annual Meeting of the Association
  for Computational Linguistics (Volume 1: Long Papers)}, pages 1657--1668,
  Vancouver, Canada. Association for Computational Linguistics.

\bibitem[{Cross and Huang(2016)}]{cross-huang-2016-span}
James Cross and Liang Huang. 2016.
\newblock \href {https://doi.org/10.18653/v1/D16-1001} {Span-based constituency
  parsing with a structure-label system and provably optimal dynamic oracles}.
\newblock In \emph{Proceedings of the 2016 Conference on Empirical Methods in
  Natural Language Processing}, pages 1--11, Austin, Texas. Association for
  Computational Linguistics.

\bibitem[{Cui and Zhang(2019)}]{lan}
Leyang Cui and Yue Zhang. 2019.
\newblock \href {https://doi.org/10.18653/v1/D19-1422} {Hierarchically-refined
  label attention network for sequence labeling}.
\newblock In \emph{Proceedings of the 2019 Conference on Empirical Methods in
  Natural Language Processing and the 9th International Joint Conference on
  Natural Language Processing (EMNLP-IJCNLP)}, pages 4115--4128, Hong Kong,
  China. Association for Computational Linguistics.

\bibitem[{Dinan et~al.(2019)Dinan, Roller, Shuster, Fan, Auli, and
  Weston}]{wizard}
Emily Dinan, Stephen Roller, Kurt Shuster, Angela Fan, Michael Auli, and Jason
  Weston. 2019.
\newblock \href {https://openreview.net/forum?id=r1l73iRqKm} {Wizard of
  wikipedia: Knowledge-powered conversational agents}.
\newblock In \emph{International Conference on Learning Representations}.

\bibitem[{Durrett and Klein(2015)}]{crf-parsing}
Greg Durrett and Dan Klein. 2015.
\newblock \href {https://doi.org/10.3115/v1/P15-1030} {Neural {CRF} parsing}.
\newblock In \emph{Proceedings of the 53rd Annual Meeting of the Association
  for Computational Linguistics and the 7th International Joint Conference on
  Natural Language Processing (Volume 1: Long Papers)}, pages 302--312,
  Beijing, China. Association for Computational Linguistics.

\bibitem[{Fried et~al.(2019)Fried, Kitaev, and Klein}]{cross-domain-parser}
Daniel Fried, Nikita Kitaev, and Dan Klein. 2019.
\newblock \href {https://doi.org/10.18653/v1/P19-1031} {Cross-domain
  generalization of neural constituency parsers}.
\newblock In \emph{Proceedings of the 57th Annual Meeting of the Association
  for Computational Linguistics}, pages 323--330, Florence, Italy. Association
  for Computational Linguistics.

\bibitem[{Gaddy et~al.(2018)Gaddy, Stern, and Klein}]{going-on-parser}
David Gaddy, Mitchell Stern, and Dan Klein. 2018.
\newblock \href {https://doi.org/10.18653/v1/N18-1091} {What{'}s going on in
  neural constituency parsers? an analysis}.
\newblock In \emph{Proceedings of the 2018 Conference of the North {A}merican
  Chapter of the Association for Computational Linguistics: Human Language
  Technologies, Volume 1 (Long Papers)}, pages 999--1010, New Orleans,
  Louisiana. Association for Computational Linguistics.

\bibitem[{Goldberg and Nivre(2013)}]{error-propagation}
Yoav Goldberg and Joakim Nivre. 2013.
\newblock \href {https://doi.org/10.1162/tacl_a_00237} {Training deterministic
  parsers with non-deterministic oracles}.
\newblock \emph{Transactions of the Association for Computational Linguistics},
  1:403--414.

\bibitem[{Gui et~al.(2020)Gui, Ye, Zhang, Li, Fei, Gong, and
  Huang}]{gui-etal-2020-uncertainty}
Tao Gui, Jiacheng Ye, Qi~Zhang, Zhengyan Li, Zichu Fei, Yeyun Gong, and
  Xuanjing Huang. 2020.
\newblock \href {https://doi.org/10.18653/v1/2020.emnlp-main.181}
  {Uncertainty-aware label refinement for sequence labeling}.
\newblock In \emph{Proceedings of the 2020 Conference on Empirical Methods in
  Natural Language Processing (EMNLP)}, pages 2316--2326, Online. Association
  for Computational Linguistics.

\bibitem[{He and McAuley(2016)}]{review}
Ruining He and Julian McAuley. 2016.
\newblock \href {https://doi.org/10.1145/2872427.2883037} {Ups and downs:
  Modeling the visual evolution of fashion trends with one-class collaborative
  filtering}.
\newblock In \emph{Proceedings of the 25th International Conference on World
  Wide Web}, WWW '16, pages 507--517, Republic and Canton of Geneva,
  Switzerland. International World Wide Web Conferences Steering Committee.

\bibitem[{Kitaev et~al.(2019)Kitaev, Cao, and
  Klein}]{kitaev-etal-2019-multi-lingual}
Nikita Kitaev, Steven Cao, and Dan Klein. 2019.
\newblock \href {https://doi.org/10.18653/v1/P19-1340} {Multilingual
  constituency parsing with self-attention and pre-training}.
\newblock In \emph{Proceedings of the 57th Annual Meeting of the Association
  for Computational Linguistics}, pages 3499--3505, Florence, Italy.
  Association for Computational Linguistics.

\bibitem[{Kitaev and Klein(2018)}]{san-constituency}
Nikita Kitaev and Dan Klein. 2018.
\newblock \href {https://doi.org/10.18653/v1/P18-1249} {Constituency parsing
  with a self-attentive encoder}.
\newblock In \emph{Proceedings of the 56th Annual Meeting of the Association
  for Computational Linguistics (Volume 1: Long Papers)}, pages 2676--2686,
  Melbourne, Australia. Association for Computational Linguistics.

\bibitem[{Kummerfeld et~al.(2012)Kummerfeld, Hall, Curran, and
  Klein}]{kummerfeld-etal-2012-parser}
Jonathan~K. Kummerfeld, David Hall, James~R. Curran, and Dan Klein. 2012.
\newblock \href {https://aclanthology.org/D12-1096} {Parser showdown at the
  {W}all {S}treet corral: An empirical investigation of error types in parser
  output}.
\newblock In \emph{Proceedings of the 2012 Joint Conference on Empirical
  Methods in Natural Language Processing and Computational Natural Language
  Learning}, pages 1048--1059, Jeju Island, Korea. Association for
  Computational Linguistics.

\bibitem[{Liu and Zhang(2017)}]{liu-zhang-2017-order}
Jiangming Liu and Yue Zhang. 2017.
\newblock \href {https://doi.org/10.1162/tacl_a_00070} {In-order
  transition-based constituent parsing}.
\newblock \emph{Transactions of the Association for Computational Linguistics},
  5:413--424.

\bibitem[{Ma and Hovy(2016)}]{lstm-crf}
Xuezhe Ma and Eduard~H. Hovy. 2016.
\newblock \href {http://arxiv.org/abs/1603.01354} {End-to-end sequence labeling
  via bi-directional lstm-cnns-crf}.
\newblock \emph{CoRR}, abs/1603.01354.

\bibitem[{Marcus et~al.(1993)Marcus, Santorini, and Marcinkiewicz}]{ptb}
Mitchell~P. Marcus, Beatrice Santorini, and Mary~Ann Marcinkiewicz. 1993.
\newblock \href {https://www.aclweb.org/anthology/J93-2004} {Building a large
  annotated corpus of {E}nglish: The {P}enn {T}reebank}.
\newblock \emph{Computational Linguistics}, 19(2):313--330.

\bibitem[{McDonald and Nivre(2011)}]{analyzing-dependency}
Ryan McDonald and Joakim Nivre. 2011.
\newblock \href {https://doi.org/10.1162/coli_a_00039} {Analyzing and
  integrating dependency parsers}.
\newblock \emph{Computational Linguistics}, 37(1):197--230.

\bibitem[{Mrini et~al.(2020)Mrini, Dernoncourt, Tran, Bui, Chang, and
  Nakashole}]{label-attention-parsing}
Khalil Mrini, Franck Dernoncourt, Quan~Hung Tran, Trung Bui, Walter Chang, and
  Ndapa Nakashole. 2020.
\newblock \href {https://doi.org/10.18653/v1/2020.findings-emnlp.65}
  {Rethinking self-attention: Towards interpretability in neural parsing}.
\newblock In \emph{Findings of the Association for Computational Linguistics:
  EMNLP 2020}, pages 731--742, Online. Association for Computational
  Linguistics.

\bibitem[{Nguyen et~al.(2020)Nguyen, Nguyen, Joty, and Li}]{constituency-point}
Thanh-Tung Nguyen, Xuan-Phi Nguyen, Shafiq Joty, and Xiaoli Li. 2020.
\newblock \href {https://doi.org/10.18653/v1/2020.acl-main.301} {Efficient
  constituency parsing by pointing}.
\newblock In \emph{Proceedings of the 58th Annual Meeting of the Association
  for Computational Linguistics}, pages 3284--3294, Online. Association for
  Computational Linguistics.

\bibitem[{Petrov and Klein(2007)}]{em}
Slav Petrov and Dan Klein. 2007.
\newblock \href {https://aclanthology.org/N07-1051} {Improved inference for
  unlexicalized parsing}.
\newblock In \emph{Human Language Technologies 2007: The Conference of the
  North {A}merican Chapter of the Association for Computational Linguistics;
  Proceedings of the Main Conference}, pages 404--411, Rochester, New York.
  Association for Computational Linguistics.

\bibitem[{Pislar and Rei(2020)}]{forest-ner}
Miruna Pislar and Marek Rei. 2020.
\newblock \href {https://doi.org/10.18653/v1/2020.coling-main.335} {Seeing both
  the forest and the trees: Multi-head attention for joint classification on
  different compositional levels}.
\newblock In \emph{Proceedings of the 28th International Conference on
  Computational Linguistics}, pages 3761--3775, Barcelona, Spain (Online).
  International Committee on Computational Linguistics.

\bibitem[{Seddah et~al.(2013)Seddah, Tsarfaty, K{\"u}bler, Candito, Choi,
  Farkas, Foster, Goenaga, Gojenola~Galletebeitia, Goldberg, Green, Habash,
  Kuhlmann, Maier, Nivre, Przepi{\'o}rkowski, Roth, Seeker, Versley, Vincze,
  Woli{\'n}ski, Wr{\'o}blewska, and Villemonte de~la Clergerie}]{spmrl}
Djam{\'e} Seddah, Reut Tsarfaty, Sandra K{\"u}bler, Marie Candito, Jinho~D.
  Choi, Rich{\'a}rd Farkas, Jennifer Foster, Iakes Goenaga, Koldo
  Gojenola~Galletebeitia, Yoav Goldberg, Spence Green, Nizar Habash, Marco
  Kuhlmann, Wolfgang Maier, Joakim Nivre, Adam Przepi{\'o}rkowski, Ryan Roth,
  Wolfgang Seeker, Yannick Versley, Veronika Vincze, Marcin Woli{\'n}ski, Alina
  Wr{\'o}blewska, and Eric Villemonte de~la Clergerie. 2013.
\newblock \href {https://aclanthology.org/W13-4917} {Overview of the {SPMRL}
  2013 shared task: A cross-framework evaluation of parsing morphologically
  rich languages}.
\newblock In \emph{Proceedings of the Fourth Workshop on Statistical Parsing of
  Morphologically-Rich Languages}, pages 146--182, Seattle, Washington, USA.
  Association for Computational Linguistics.

\bibitem[{Stern et~al.(2017{\natexlab{a}})Stern, Andreas, and
  Klein}]{minimal-span}
Mitchell Stern, Jacob Andreas, and Dan Klein. 2017{\natexlab{a}}.
\newblock \href {https://doi.org/10.18653/v1/P17-1076} {A minimal span-based
  neural constituency parser}.
\newblock In \emph{Proceedings of the 55th Annual Meeting of the Association
  for Computational Linguistics (Volume 1: Long Papers)}, pages 818--827,
  Vancouver, Canada. Association for Computational Linguistics.

\bibitem[{Stern et~al.(2017{\natexlab{b}})Stern, Andreas, and
  Klein}]{stern-etal-2017-minimal}
Mitchell Stern, Jacob Andreas, and Dan Klein. 2017{\natexlab{b}}.
\newblock \href {https://doi.org/10.18653/v1/P17-1076} {A minimal span-based
  neural constituency parser}.
\newblock In \emph{Proceedings of the 55th Annual Meeting of the Association
  for Computational Linguistics (Volume 1: Long Papers)}, pages 818--827,
  Vancouver, Canada. Association for Computational Linguistics.

\bibitem[{Stiansen and Voeten(2019)}]{ECtHR}
Øyvind Stiansen and Erik Voeten. 2019.
\newblock \href {https://doi.org/10.7910/DVN/OBYUO5} {{ECtHR judgments}}.

\bibitem[{Teng and Zhang(2018)}]{two-local-model}
Zhiyang Teng and Yue Zhang. 2018.
\newblock \href {https://www.aclweb.org/anthology/C18-1011} {Two local models
  for neural constituent parsing}.
\newblock In \emph{Proceedings of the 27th International Conference on
  Computational Linguistics}, pages 119--132, Santa Fe, New Mexico, USA.
  Association for Computational Linguistics.

\bibitem[{Tian et~al.(2020)Tian, Song, Xia, and Zhang}]{span-attention}
Yuanhe Tian, Yan Song, Fei Xia, and Tong Zhang. 2020.
\newblock \href {https://doi.org/10.18653/v1/2020.findings-emnlp.153}
  {Improving constituency parsing with span attention}.
\newblock In \emph{Findings of the Association for Computational Linguistics:
  EMNLP 2020}, pages 1691--1703, Online. Association for Computational
  Linguistics.

\bibitem[{V{\"o}lske et~al.(2017)V{\"o}lske, Potthast, Syed, and
  Stein}]{voelske:2017}
Michael V{\"o}lske, Martin Potthast, Shahbaz Syed, and Benno Stein. 2017.
\newblock \href {https://doi.org/10.18653/v1/W17-4508} {{TL;DR: Mining Reddit
  to Learn Automatic Summarization}}.
\newblock In \emph{Workshop on New Frontiers in Summarization at EMNLP 2017},
  pages 59--63. Association for Computational Linguistics.

\bibitem[{Wang et~al.(2018)Wang, Pham, Yin, and Neubig}]{tree-mt}
Xinyi Wang, Hieu Pham, Pengcheng Yin, and Graham Neubig. 2018.
\newblock \href {https://doi.org/10.18653/v1/D18-1509} {A tree-based decoder
  for neural machine translation}.
\newblock In \emph{Proceedings of the 2018 Conference on Empirical Methods in
  Natural Language Processing}, pages 4772--4777, Brussels, Belgium.
  Association for Computational Linguistics.

\bibitem[{Xu and Durrett(2019)}]{cp-summarization}
Jiacheng Xu and Greg Durrett. 2019.
\newblock \href {https://doi.org/10.18653/v1/D19-1324} {Neural extractive text
  summarization with syntactic compression}.
\newblock In \emph{Proceedings of the 2019 Conference on Empirical Methods in
  Natural Language Processing and the 9th International Joint Conference on
  Natural Language Processing (EMNLP-IJCNLP)}, pages 3292--3303, Hong Kong,
  China. Association for Computational Linguistics.

\bibitem[{Xue et~al.(2005)Xue, Xia, Chiou, and Palmer}]{ctb}
Naiwen Xue, Fei Xia, Fu-dong Chiou, and Marta Palmer. 2005.
\newblock \href {https://doi.org/10.1017/S135132490400364X} {The penn chinese
  treebank: Phrase structure annotation of a large corpus}.
\newblock \emph{Nat. Lang. Eng.}, 11(2):207–238.

\bibitem[{Yang et~al.(2022)Yang, Cui, Ning, Wu, and
  Zhang}]{yang-etal-2022-challenge}
Sen Yang, Leyang Cui, Ruoxi Ning, Di~Wu, and Yue Zhang. 2022.
\newblock Challenges to open-domain constituency parsing.
\newblock In \emph{Findings of the Association for Computational Linguistics:
  ACL 2022}.

\bibitem[{Zhang and Zhang(2010)}]{mll}
Min-Ling Zhang and Kun Zhang. 2010.
\newblock \href {https://doi.org/10.1145/1835804.1835930} {Multi-label learning
  by exploiting label dependency}.
\newblock In \emph{Proceedings of the 16th ACM SIGKDD International Conference
  on Knowledge Discovery and Data Mining}, KDD '10, page 999–1008, New York,
  NY, USA. Association for Computing Machinery.

\bibitem[{Zhang et~al.(2020{\natexlab{a}})Zhang, Li, and
  Zhang}]{zhang-etal-2020-efficient}
Yu~Zhang, Zhenghua Li, and Min Zhang. 2020{\natexlab{a}}.
\newblock \href {https://doi.org/10.18653/v1/2020.acl-main.302} {Efficient
  second-order {T}ree{CRF} for neural dependency parsing}.
\newblock In \emph{Proceedings of the 58th Annual Meeting of the Association
  for Computational Linguistics}, pages 3295--3305, Online. Association for
  Computational Linguistics.

\bibitem[{Zhang et~al.(2020{\natexlab{b}})Zhang, Zhou, and
  Li}]{constituency-crf}
Yu~Zhang, Houquan Zhou, and Zhenghua Li. 2020{\natexlab{b}}.
\newblock \href {https://doi.org/10.24963/ijcai.2020/560} {Fast and accurate
  neural crf constituency parsing}.
\newblock In \emph{Proceedings of the Twenty-Ninth International Joint
  Conference on Artificial Intelligence, {IJCAI-20}}, pages 4046--4053.
  International Joint Conferences on Artificial Intelligence Organization.

\bibitem[{Zhang and Clark(2009)}]{zhang-clark-2009-transition}
Yue Zhang and Stephen Clark. 2009.
\newblock \href {https://aclanthology.org/W09-3825} {Transition-based parsing
  of the {C}hinese treebank using a global discriminative model}.
\newblock In \emph{Proceedings of the 11th International Conference on Parsing
  Technologies ({IWPT}{'}09)}, pages 162--171, Paris, France. Association for
  Computational Linguistics.

\bibitem[{Zhang and Nivre(2012)}]{zhang-nivre-2012-analyzing}
Yue Zhang and Joakim Nivre. 2012.
\newblock \href {https://aclanthology.org/C12-2136} {Analyzing the effect of
  global learning and beam-search on transition-based dependency parsing}.
\newblock In \emph{Proceedings of {COLING} 2012: Posters}, pages 1391--1400,
  Mumbai, India. The COLING 2012 Organizing Committee.

\bibitem[{Zhou and Xu(2015)}]{zhou-xu-2015-end}
Jie Zhou and Wei Xu. 2015.
\newblock \href {https://doi.org/10.3115/v1/P15-1109} {End-to-end learning of
  semantic role labeling using recurrent neural networks}.
\newblock In \emph{Proceedings of the 53rd Annual Meeting of the Association
  for Computational Linguistics and the 7th International Joint Conference on
  Natural Language Processing (Volume 1: Long Papers)}, pages 1127--1137,
  Beijing, China. Association for Computational Linguistics.

\bibitem[{Zhou and Zhao(2019)}]{head-driven}
Junru Zhou and Hai Zhao. 2019.
\newblock \href {https://doi.org/10.18653/v1/P19-1230} {{H}ead-{D}riven
  {P}hrase {S}tructure {G}rammar parsing on {P}enn {T}reebank}.
\newblock In \emph{Proceedings of the 57th Annual Meeting of the Association
  for Computational Linguistics}, pages 2396--2408, Florence, Italy.
  Association for Computational Linguistics.

\end{thebibliography}


\end{document}